\documentclass[letterpaper]{article} 
\usepackage{aaai23}  
\usepackage{times}  
\usepackage{helvet}  
\usepackage{courier}  
\usepackage[hyphens]{url}  
\usepackage{graphicx} 
\usepackage{natbib}  
\usepackage{caption} 
\usepackage{algorithm}
\usepackage{algorithmic}

\usepackage{enumitem}

\usepackage{newfloat}
\usepackage{listings}
\DeclareCaptionStyle{ruled}{labelfont=normalfont,labelsep=colon,strut=off} 
\floatstyle{ruled}
\newfloat{listing}{tb}{lst}{}
\floatname{listing}{Listing}
\usepackage{booktabs}
\usepackage{amsmath}
\usepackage{syntax}

\title{A Neurodiversity-Inspired Solver for the Abstraction \& Reasoning Corpus (ARC) Using Visual Imagery and Program Synthesis}
\author {
    James Ainooson, 
    Deepayan Sanyal, 
    Joel P. Michelson, 
    Yuan Yang,
    Maithilee Kunda 
  }

\affiliations {
    Department of Computer Science, Vanderbilt University.\\
    james.ainooson@vanderbilt.edu, deepayan.sanyal@vanderbilt.edu, joel.p.michelson@vanderbilt.edu, yuan.yang@vanderbilt.edu, mkunda@vanderbilt.edu
}

\begin{document}

\maketitle
\begin{abstract}
Core knowledge about physical objects---e.g., their permanency, spatial transformations, and interactions---is one of the most fundamental building blocks of biological intelligence across humans and non-human animals.  While AI techniques in certain domains (e.g. vision, NLP) have advanced dramatically in recent years, no current AI systems can yet match human abilities in flexibly applying core knowledge to solve novel tasks.  We propose a new AI approach to core knowledge that combines 1) visual representations of core knowledge inspired by human mental imagery abilities, especially as observed in studies of neurodivergent individuals; with 2) tree-search-based program synthesis for flexibly combining core knowledge to form new reasoning strategies on the fly.  We demonstrate our system's performance on the very difficult Abstraction \& Reasoning Corpus (ARC) challenge, and we share experimental results from publicly available ARC items as well as from our 4th-place finish on the private test set during the 2022 global ARCathon challenge.
\end{abstract}

	\section{Introduction}

\begin{figure}[t]
	\includegraphics[width=\linewidth]{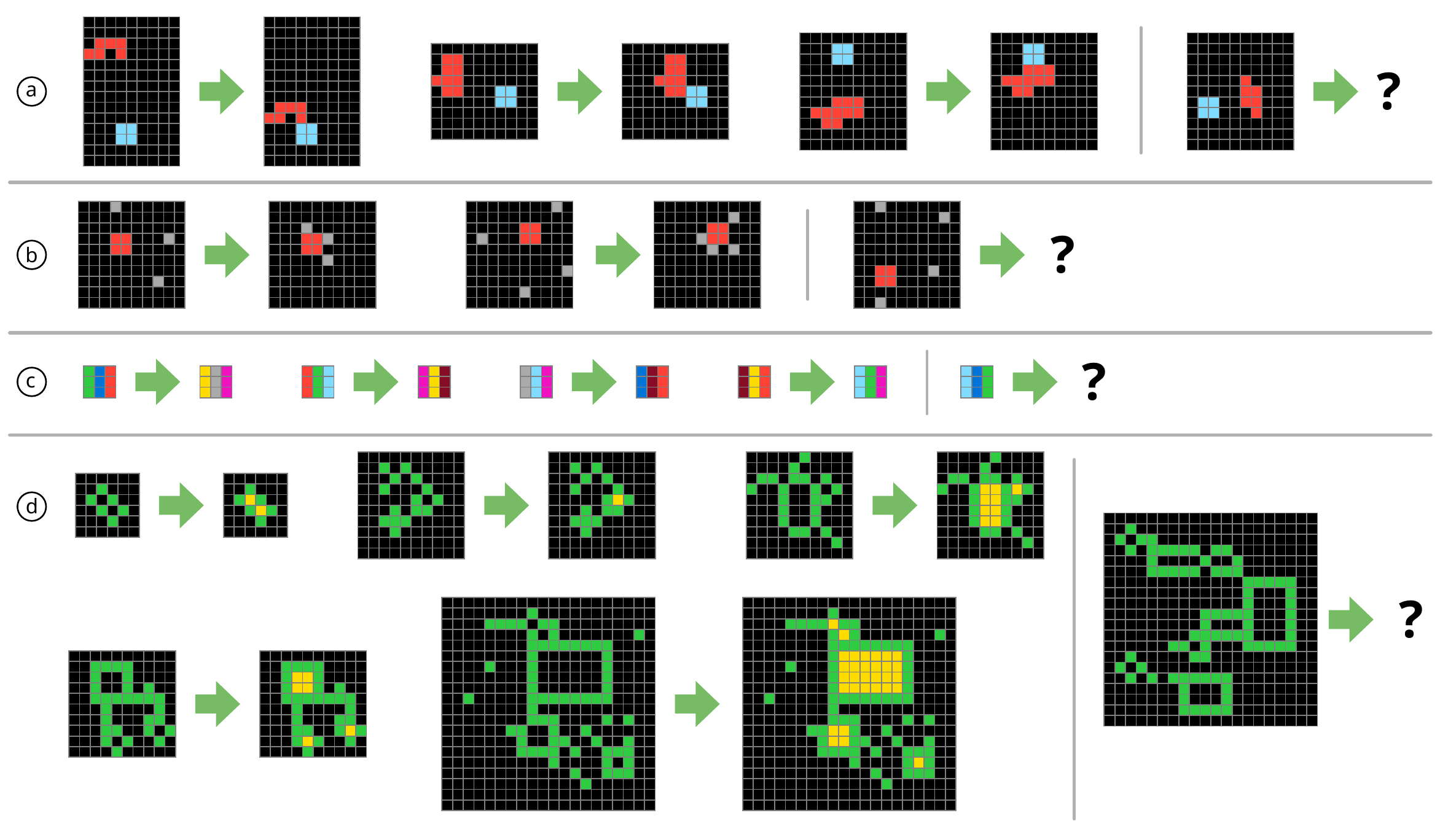}
	\caption{Four sample tasks from the Abstraction \& Reasoning Corpus (ARC).  Each task consists of 1-5 input-output ``training'' pairs plus 1-2 input-only ``test'' pairs.}
	\label{arc-task-2}
\end{figure}

Consider the four visual reasoning problems depicted in the rows of Figure \ref{arc-task-2}.  Each problem presents you, the problem-solver, with a small number of input-output image pairs on the left, and then you must provide the correct output for the final input image on the right.

Most people can likely solve many of these problems without much difficulty.  The knowledge being tested might seem fairly intuitive and easy to describe, e.g., for the first item, the red object gets attracted to the blue block.  For the second, the ``stones'' get attracted to the red block.  The third involves color-swapping.  For the fourth, we fill in all of the closed holes with yellow.

What is this knowledge, and how do we know it and use it?  For these problems, the required knowledge has to do with the physical nature of objects in our world, and how this physical nature appears to our visual sensory system.  For instance, we know and can reason about rigid objects with coherent boundaries; motions of these objects in various directions; contacts between multiple objects; changes in object colors while retaining the same shape; etc.

In the cognitive sciences, this type of knowledge has been called \textit{core knowledge about objects}, and it is believed to be one of the most fundamental building blocks of biological intelligence across both humans and many non-human animals, alongside other core knowledge systems for reasoning about actions, number, space, and social agents \cite{spelke2007core}.  While AI techniques in certain domains (e.g. vision, NLP) have advanced dramatically in recent years, no current AI systems can yet match human abilities in flexibly applying core knowledge to solve novel tasks.  Moreover, because core knowledge about objects feeds into virtually every other domain of reasoning, from spatial to quantitative to linguistic to social, this gap will inevitably lead to upper bounds on the robustness and flexibility of current AI reasoning systems across all of these domains.

Why is core knowledge such a tough nut for AI to crack?  One challenge is that, despite extensive evidence that core knowledge systems exist in biological intelligence, cognitive scientists have relatively little idea about how core knowledge is represented, and scientific debates continue to rage about how core knowledge is acquired and deployed \cite[e.g.,][]{smith2020beyond}.

In this paper, we propose a new formulation for visually representing and reasoning about core knowledge:

\begin{enumerate}[nolistsep,noitemsep]
	\item We describe an imperative-style domain specific language, called the Visual Imagery Reasoning Language (VIMRL), that represents core knowledge as a collection of image-transformation functions that are inspired by cognitive studies of human visual mental imagery, including studies of neurodivergent forms of imagery-based reasoning in individuals on the autism spectrum. 
	\item We propose an innovative reasoning method that combines VIMRL with multi-level search-based program synthesis to flexibly select, compose, and deploy core knowledge on novel tasks.  
	\item  We demonstrate our system's performance on the very difficult Abstraction \& Reasoning Corpus (ARC) challenge \cite{Chollet2019}, and we share experimental results from publicly available ARC items as well as from our 4th-place finish on the private test set during the 2022 global ARCathon challenge.
\end{enumerate}

\section{The Abstraction \& Reasoning Corpus}

 \citet{Chollet2019} recently introduced the Abstraction \& Reasoning Corpus (ARC) as a general intelligence test for humans and artificial systems designed to load on four types of core knowledge priors.  The \textbf{objectness} prior requires solvers to deal with the segmentation, permanence and interaction of objects. Tasks that require \textbf{goal-directedness} may exhibit input-output grids that can be considered as the start and end states of some abstract process, such as an object moving from one point in the grid to another. 
\textbf{Numbers and counting} priors are required in situations where quantities, like frequencies of object occurrences and sizes of objects, serve to influence operations like comparison and sorting. 
The \textbf{geometry and topology} prior requires an agent to have knowledge about shapes, lines, symmetry, relative positioning of objects, and the ability to replicate objects in different ways.

Fig. \ref{arc-task-2} shows four sample ARC tasks. Each of these tasks has a training section (left of the grey bar), with a number of input-output grid pairs (usually 1-5), and a test section (right of the grey bar) with a number of input-only grids (usually 1-2) for which the solver must provide an output. 


All tasks from the ARC are presented on coloured, input-output grid pairs. These grids can be anywhere from a single cell to 30x30 cells in size, and each cell can contain one of ten symbols (0 through 9 or appropriately selected unique colours). A property that makes the ARC particularly difficult, especially for artificial systems, is how solvers are required to predict the output grids from scratch: solvers must correctly determine the grid's size, and must also specify the symbols to place in each cell correctly.


The ARC contains a 1,000 different tasks, divided into a publicly available ``training set'' of 400 tasks, a publicly available ``evaluation'' set of 400 tasks, and a private test set of 200 tasks accessible through submissions to various ARC challenge platforms (e.g., Kaggle, recent ARCathon challenges). 
To help researchers develop and test their solvers, the complete solutions (containing both the input and output pairs) are supplied for all 800 publicly available tasks.  

Interestingly, while the public task sets are labeled ``training'' and ``evaluation,'' these do not comprise training and evaluation sets in the traditional machine learning sense, as according to \cite{Chollet2019}, the concepts observed in one set do not transfer to the other. This means, agents that specialize on items in one set may not necessarily perform well on items in the other.  This pattern also seems to hold across the private test set as well, making the ARC an extremely difficult test for current AI systems. 


Formally an ARC task, \( T \), can be defined as: \[T = \Bigl\{ \langle I^t_1, O^t_1 \rangle, \ldots, \langle I^t_n, O^t_n \rangle;  I^e_0 , \dots , O^e_m \Bigr\}\] Here, \(I^t\) and \(O^t\) are input and output training grids, and \(i^e\) represents a test input grid for which the agent must provide an output. In the case of the publicly  available ARC tasks, each test input, \(I^e_i\), has a corresponding output, \(O^e_i\). 

Scoring on the ARC is considered an all or nothing affair: solvers must get all the cells right to correctly solve a task. Any missed cell results in a failure. To allow some room for error, solvers are given the opportunity to make up to three predictions, and the task will be considered solved if any of those predictions are correct.



\section{Core Knowledge, Visual Imagery, and Autism}

How might we formalize representations of core knowledge in a computational system?  Our work takes inspiration from studies of human reasoning, and in particular studies of neurodivergent reasoning in individuals on the autism spectrum.

Raven's Progressive Matrices (RPM) is a very well known and widely used human intelligence test that is similar to the ARC in flavor, though much smaller in scope, with 60 problems that cover just a couple of dozen core knowledge concepts in total.  Though the RPM does not explicitly load on language, many early studies spanning psychology, neuroscience, and AI emphasized the role of symbolic and/or linguistic representations in human RPM problem solving \cite[e.g.,][]{carpenter1990one,prabhakaran1997neural}, especially on harder test problems.  In other words, it was assumed that perceptual inputs were processed to create amodal representations, and then core knowledge functions could operate over these amodal representations using symbolic, linguistic, and/or qualitative operations.

However, unusual (and strong) patterns of RPM performance were later documented in individuals on the autism spectrum \cite{dawson2007level}.  These patterns were linked to atypically strong uses of visual imagery in these individuals \cite{soulieres2009enhanced} which also matched introspective accounts of reasoning \cite[e.g.,][]{grandin2008thinking}.  Visual imagery refers to an agent's use of internal representations that are retinotopic, i.e., that share rich structural correspondence with visual perceptual representations but that are not directly tied to perceptual inputs \cite{pearson2015heterogeneity}.

Computational, imagery-based RPM models followed suit and were successful \cite{kunda2013computational,mcgreggor2014fractals,yang2020not}.  In these models, core knowledge concepts are represented as image transformation operations that can be applied directly to perceptual inputs, without requiring separate stages of feature or object extraction.

We take inspiration from these convergent studies of the effectiveness of visual-imagery-based representations on core knowledge tasks.  In particular, our approach is based around a new Visual Imagery Reasoning Language (VIMRL) that explicitly represents core knowledge priors as visual-imagery-like functions.


\section{Our Approach: VIMRL + Program Synthesis}\label{reasoning}

While the Visual Imagery Reasoning Language (VIMRL) framework is intended to be generalizable to core knowledge about objects in varied contexts, we focus this paper on a version of VIMRL specialized for application to ARC-style image inputs and outputs, along with a program synthesis solver that generates VIMRL programs. 
When given an ARC task, \(T=\Bigl\{ \langle I^t_1, O^t_1 \rangle, \ldots, \langle I^t_n, O^t_n \rangle;  I^e_0 , \dots , I^e_m \Bigr\}\), the solver searches the space of programs in VIMRL for a candidate program, \( \varphi (x) \), which takes a grid, $x$, as input, produces a solution output grid. For a program to be considered as a candidate solution, it must satisfy $\left(\frac{\sum_{i=1}^{n} \lambda(\varphi(I^t_i), O^t_i)}{n} > \alpha \right)$, where $ \lambda(x, y) = \begin{cases}
	1 & x = y \\
	0
\end{cases} $ and, $\alpha$, is a threshold within which the solver should be accurate on training items. This means a candidate program will only be selected if it solves enough items in a task's training problems with an accuracy higher than the value of $\alpha$.

\subsection{Visual Imagery Reasoning Language}
The Visual Imagery Reasoning Language (VIMRL) is an imperative style language, designed around imagery operations, and built specifically for reasoning about ARC tasks. Instead of relying on control instructions, VIMRL places emphasis on the sequence of instructions to control the state of a program during execution. See Table \ref{fig:language-grammar} for VIMRL's full grammar.

\grammarindent0.8in
\begin{table}[b]
	\begin{tabular}{p{3.2in}}
		\toprule
		\begin{grammar}
			<instruction> ::= <assignment> 
			\alt <operation>
		\end{grammar}\vspace{-7mm} \\ \midrule\begin{grammar}
			<assignment> ::= <identifier> `='  <operation> 
		\end{grammar}\vspace{-7mm} \\ \midrule\begin{grammar}
			<operation> ::= <identifier> `(' <arguments> `)'
		\end{grammar}\vspace{-7mm} \\ \midrule\begin{grammar}
			<arguments> ::= <argument>
			\alt <arguments> `,' <argument>
		\end{grammar}\vspace{-7mm} \\ \midrule\begin{grammar}
			<argument> ::= <identifier>
			\alt <number>
			\alt <operation>
		\end{grammar}\vspace{-7mm} \\ \midrule\begin{grammar}
			<number> ::= $(`-')?[0-9]+$
		\end{grammar}\vspace{-7mm} \\ \midrule\begin{grammar}
			<identifier> ::= $[a-zA-Z][a-zA-Z0-9]*$
			
		\end{grammar} \vspace{-7mm} \\
		\bottomrule
	\end{tabular}	
	\caption{Grammar for the Visual Imagery Reasoning Language-I (VIMRL). These also double as production rules for generating code during program synthesis.}
	\label{fig:language-grammar}
\end{table}

Every instruction in the VIMRL involves a call to an operation. Operations can take arguments, which are either literal values or references to variables. And operations always return values. Values in VIMRL can either be literal or variable, and values always have fixed types.


All values (literal or variable) in VIMRL have fixed types. Arguments for operations are also expected to have specific types. Currently, values can assume one of 5 given types. Values can be typed as \texttt{image}, \texttt{object} (an image fragment with a location, much like a sprite in a video game), \texttt{color}, \texttt{number} (integers only), or \texttt{list} (of objects).

\subsection{Executing VIMRL Programs}
A VIMRL program under execution can be considered to have a state containing the following:
\begin{enumerate} 
	\item The set of all variables that have been defined throughout the program's lifetime.
	\item All the values associated with the defined variables.
	\item The current line of instruction being executed.
\end{enumerate}

Every program starts execution with two pre-defined variables: \texttt{input}, an image containing the input grid of the task and \texttt{background}, a value fors the background colour of the input grid. The background value helps in isolating objects, and by default it is set to a value of \texttt{0}. This default value can, however, be changed through other operations that are executed later.


During execution, two main types of operations can be performed. First, there are low level operations, which are simple functions (like \texttt{trim}) that require arguments to be explicitly passed. These operations will typically manipulate their inputs in a definite way and return an output. The second type of operations are high level operations, which take a single argument and further analyse the grids from the tasks to be solved, and the current state of execution to implicitly select extra arguments. 

\begin{table}
	\begin{tabular}{l|l}
		\toprule
		(a) & \begin{minipage}{3in}
			\begin{verbatim}output = attract(input)\end{verbatim}
		\end{minipage}\\
		\midrule
		(b) & \begin{minipage}{3in}
			\begin{verbatim}output = recolor(input)\end{verbatim}
		\end{minipage} \\
		\midrule
		(c) & \begin{minipage}{3in}
			\begin{verbatim}enclosed = find_enclosed_patches(input)
				recolored = recolor_objects(enclosed)
				output = draw(input, recolored)
			\end{verbatim}
		\end{minipage}\\
		\bottomrule
	\end{tabular}
	\caption {Three sample VIMRL programs that solve ARC tasks. Cell (a) provides possible solutions for the tasks in Figures \ref{arc-task-2}(a) and \ref{arc-task-2}(b), and cells (c) and (d) provide possible solutions for the tasks in Figures \ref{arc-task-2}(c) and \ref{arc-task-2}(d), respectively.} 
	\label{tab:listings1}
	
\end{table}

To further explain how high-level operations work, consider the programs listed in Table \ref{tab:listings1}. 
The program in cell (b) can solve both of the tasks displayed in Figures \ref{arc-task-2} (a) and \ref{arc-task-2}(b). The \texttt{attract} operation used in the program is a high level operation that uses simple naive physics simulations to solve the problem of objects being attracted to each other. It takes an input image of the initial state of the objects, and returns an output image with the final state. 

Because \texttt{attract} is a high level function, all of the tasks training items, $\Bigl\{ \langle I^t_1, O^t_1 \rangle, \ldots, \langle I^t_n, O^t_n \rangle \Bigr\}$, are available to it at runtime. From these items, the \texttt{attract} function is able to form a rule about which objects are being attracted to what, and it can use this rule to attempt the test item. All high level functions in VIMRL operate this way, albeit each with their own internal rules and search techniques.

So far, all sample programs we discussed contain a single instruction that operates on the input. But programs in VIMRL are typically longer. Having multiple instructions, however, complicates execution when high level operations are performed after low level ones. Because every execution of a high level function requires an instance of the task's training items to be analysed, in cases where other instructions have already modified the task's input image, $I^e$, the input-output pairs from the training items, $\Bigl\{ \langle I^t_1, O^t_1 \rangle, \ldots, \langle I^t_n, O^t_n \rangle \Bigr\}$, may no longer be representative of the current state of the input image. 

As a solution to this problem, before any high level operation is executed, all instructions that have already been executed, are applied to all the input-output training pairs to create a modified version of the task. To do this we consider the sequence of all instructions performed before the high-level operation was executed as a partial program, $\varphi'(x)$, then we generate a modified task, $T'$, with training items, $\Bigl\{ \langle I'^t_1, O'^t_1 \rangle, \ldots, \langle I'^t_n, O'^t_n \rangle \Bigr\}$, such that $\forall_{i \in \{1,\ldots,n\}}\langle I'^t_i, O'^t_i \rangle \rightarrow \langle \varphi'(I^t_i), \varphi'(O^t_i) \rangle$. This modified task, $T'$, is then passed to the high level operation.

\subsubsection{A walk-through of high level function execution}
To further illustrate how partial programs modify the tasks before they are analysed by high level functions, consider the programs from cells (c) and (d) of Table \ref{tab:listings1}. The program in cell (c) is similar to all those we have seen earlier. It uses a single call to a high level function, \texttt{recolor}, which learns the rules by which colours in an image are transformed to solve the task in Figure \ref{arc-task-2} (c). 

\begin{figure}[h!]
	\includegraphics[width=\linewidth]{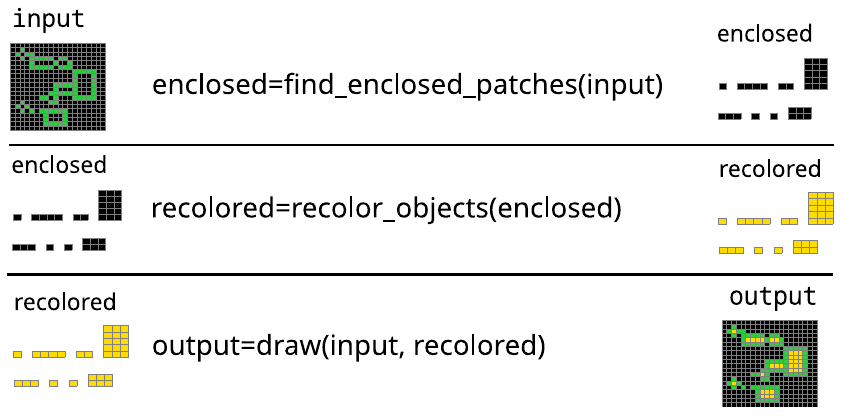}
	\caption{A walk-through of the execution of the program listed in cell (d) of Table \ref{tab:listings1} on the task from Figure \ref{arc-task-2} (d). White boxes show the state of the program's variables whiles the grey boxes show the instructions executed.}
 \label{walkthrough}
\end{figure}

When a similar \texttt{recolor} operation (\texttt{recolor_objects} which works on a list of images instead of a single image) is encountered in the program from cell (d), however, another operation, \texttt{find_enclosed_patches} has already been executed. Figure \ref{walkthrough} provides a walk-through with a visualization of the execution state as the program in cell (d) is runs.

From the walk-through we observe that the initial state has the \texttt{input} variable assigned to the input grid of the test item. The first instruction, \texttt{enclosed = find_enclosed_patches(input)}, analyzes the input image and extracts all patches of the grid's image that are enclosed. In the case of this particular example, the call to \texttt{find_enclosed_patches} returns a list with 8 grids and their locations.

The next instruction, \texttt{recolor_objects}, which is a high-level function, takes this list of objects as an argument. In addition to this list, the \texttt{recolor_objects} function will also receive a copy of the task, and this copy must reflect any changes the earlier call to \texttt{find_enclosed_patches} may have made to the input image. To build this modified task, the partial program, which contains only single a call to \texttt{find_enclosed_patches}, is executed on all the inputs and output images of the task. Whenever this partial program is executed on an image from the training set, the corresponding image in the task is replaced with the value associated with the \texttt{enclosed} variable after the partial execution. The \texttt{enclosed} variable is chosen as the replacement because it is the variable whose value is being passed to the \texttt{recolor} operation.

It is worth noting that the in this case the value of \texttt{enclosed} will be of a \texttt{list} type, leading to a situation which forces all the training images to be replaced with lists. Because the \texttt{recolor_object} operation operates on lists, it is now able to compare the lists of patches found in the input and outputs of the modified tasks to detect that everything that is coloured black (0) is switched to yellow (4).

After the execution of the \texttt{recolor} operation, the \texttt{draw} operation is used to paint all the recolored objects back to the \texttt{input} grid, and the results are assigned to the \texttt{output} variable as a solution to the task.

\subsection{Operations for Reasoning about the ARC}

\begin{figure*} [t]
	\includegraphics[width=\textwidth]{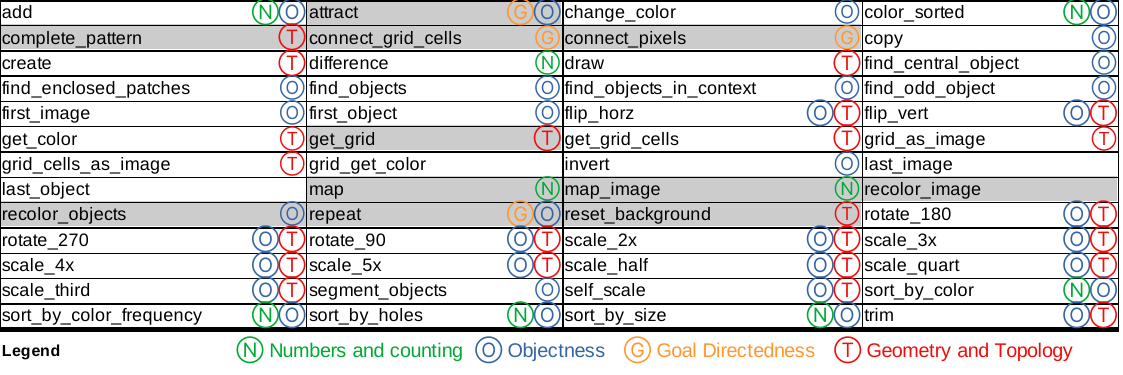}
	\caption{A sample of the instructions available for solving tasks in VIMRL. Functions listed in grey cells are high level functions, and the colored letters provide information about the core knowledge prior from which the function draws.}
	\label{operations}
\end{figure*}

Currently, there are a total of 25 high level and 51 low-level operations, giving a total of 76 operations. The implementation of these operations are inspired by the core knowledge priors suggested for the ARC and similarities observed in tasks from the public ARC datasets. A sample of these operations and the core priors upon which they are based are displayed in Figure \ref{operations}.

In addition to these 76 operations, a few extra operations are created at run-time to handle situations where functions that take single image grids as input and equally return a single image grid are modified to work on a collection of objects. These operations are much akin to the map operation typically found in functional programming languages.

It is worth noting that all operations designed for VIMRL were only made after observing just the 400 tasks in the train section of the public ARC dataset. We decided not to consider any tasks from the evaluation section in order to further evaluate how well the concepts between the tasks are separated.

\subsection{Searching}
At the core of our reasoning agent is a search algorithm that attempts to find VIMRL programs for ARC tasks. As described at the beginning of this section, the goal of the search is to find programs that solve a given number of items in from the train section of an ARC task. The major issues we had to deal with in building our search algorithm stemmed from tree traversal, successor generation, and node pruning.

Our current search algorithm can be described as follows: Given an instance of the ARC task, \(T=\Bigl\{ \langle I^t_1, O^t_1 \rangle, \ldots, \langle I^t_n, O^t_n \rangle;  I^e_0 , \dots , I^e_m \Bigr\}\), generate and collect candidate programs, \( \varphi_i (x) \), which satisfy $\left(\frac{\sum_{i=1}^{n} \lambda(\varphi(I^t_i), O^t_i)}{n} > \alpha\right) $, where $ \lambda(x, y) = \begin{cases}
	1 & x = y \\
	0
\end{cases}$, for some threshold $\alpha > 0$. The search executes in a \textit{generate-execute-test} cycle until a given number of programs (200 in the case of our experiments) are found, or a time-out (700 seconds in the case of our experiments) is reached. For all the experiments discussed in this paper, we accepted any program with $\alpha>0$ as a potential candidate.

After the search cycle terminates, the best performing programs are selected from a set of candidate programs according to two different approaches. The first approach considers the most efficient programs as candidates, whiles the second optimizes for performance and unique outputs. 
	
In the first approach, we collect the top three smallest programs with the highest scores (obtained after sorting all candidates with the $\alpha$ score in descending order and the program size in ascending order) as final candidates to be used as responses for the task. 
	
For the second approach, we first execute all candidate programs on the test items of the tasks and collect their output grids. We then group candidate programs, such that all those that produce the same output share a group. Within each group, we select the candidate program with the highest $\alpha$ score and smallest program size to be representative of the group. Then we finally select our final candidate responses by picking the top three representative programs with the highest scores.

\subsubsection{Tree traversal and successor generation}
In all our experiments, we search the space of VIMRL programs with traditional breadth-first and depth first search algorithms. Starting with an empty program, we rely on a successor generator to add on instructions as we build a series of potential programs that are evaluated on the task to be solved. 

Our approach involves executing a production system over the VIMRL grammar, with the grammar as productions, to produce all possible successors when a program is given. This yields a brute-force search through the entire VIMRL program space. With the potential to add over 70 operations at each step, the branching factor of the search tree is high, leading to a quick exponential explosion in search space. Nonetheless, a full brute force search serves as a good starting point to help us understand the dynamics of program generation in VIMRL.

\subsubsection{Search space pruning and optimization}
Even when potential successors are stochastically sampled, the search space is still expected to explode. Search space explosion cannot be entirely prevented, but steps can be taken to ensure only meaningful programs are explored and expanded.

One approach we used was to limit the depth of search to a fixed number of instructions per program. This step places a hard limit on the entire search space, making it finite. Other approaches we took involved detecting and eliminating logically equivalent programs that yield the same output and introducing reference gaps to ensure variables defined are used in search.

\begin{figure*}[t]
    \centering
    \includegraphics[width=\linewidth]{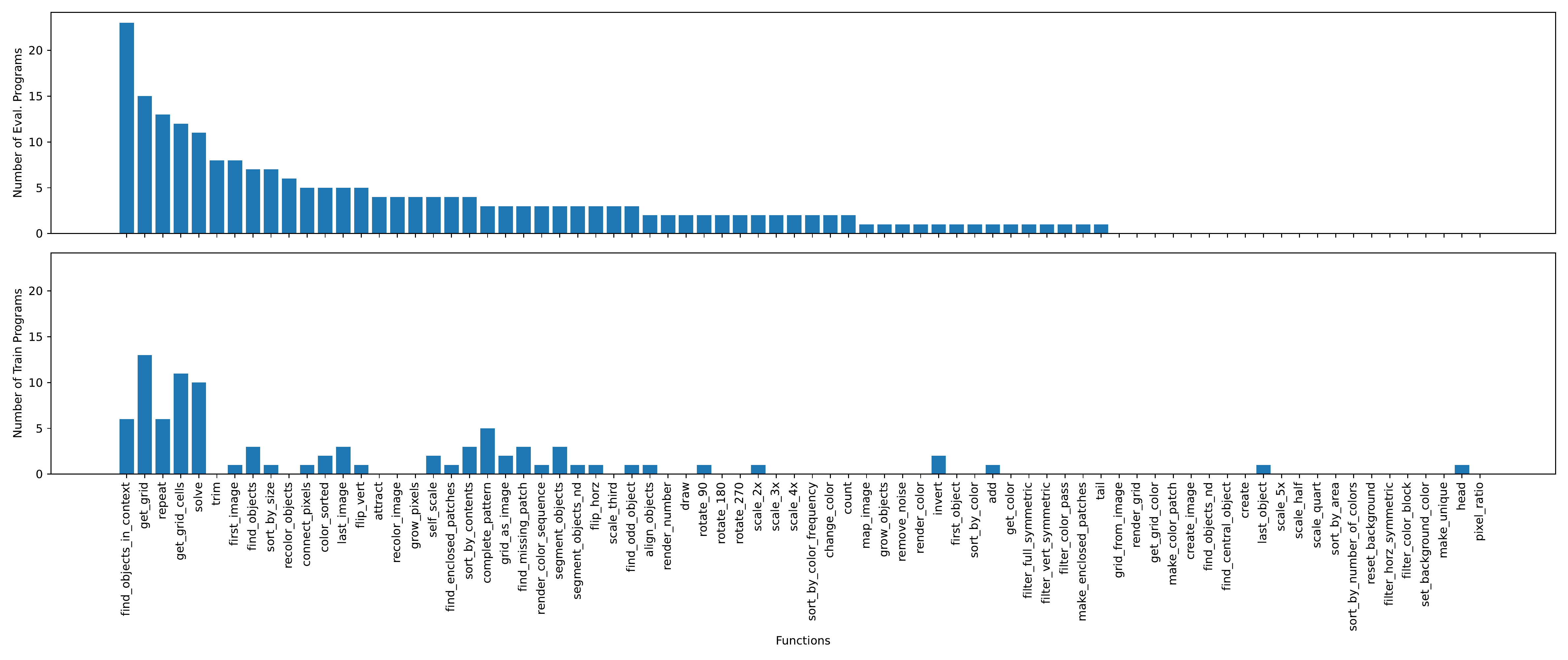}
    \caption{Distribution of operation on programs generated by our best configuration. The chart on top shows the distribution of functions used for solving items on ARC's training set, while the chart on the bottom shows the distribution of functions used for solving items on the evaluation set.}
    \label{fig:function-distribution}
\end{figure*}

\subsubsection{Logically Equivalent Programs}
There are also cases where two programs have different sequences of instructions, yet are logically equivalent (because they produce the same final program state). It is not efficient to execute logically-equivalent programs. In order to prevent logically equivalent programs from being executed, we sort all programs to ensure that logically equivalent programs have the same sequence of instructions. 

To sort these programs we build a dependency tree in which the locations of instructions that consume variables are children to the locations of instructions that create variables. When we topologically sort this dependency tree, the sequence of instruction is re-ordered such that instructions are used almost as soon as they are defined. A side effect of this re-ordering is that all programs that are logically equivalent but are sequenced differently yield the same output.

\subsubsection{Reference Gaps} 
The reference gap pruning rule ensures that variables in states are referenced within some timeframe after they are defined. This rule works by removing any nodes that have machines whose defined variables are not referenced within a given number of instructions. For variables defined in the tail end of a state's instructions, checks are made within transitions originating from the state or the list of instructions in connecting states. The number of instructions between variable definition and referencing can be set as a hyper-parameter for the search operation.

\subsubsection{Repeated Inputs}
Because inputs to programs are presented as plain variables, they are consumed during search as such. However, inputs are typically used only in the first instruction. Any subsequent reference to the input variable may likely be unnecessary. We therefore prune any such programs to improve search performance.

\subsubsection{Stochastic Search}
In earlier attempts we tried generating successors stochastically to cut down search space. Here, the probabilities of possible successor nodes were computed from hand-coded ground truth programs for a subset of the 400 training tasks of the ARC. The corpus of hand coded programs were used to build models with potential intrinsic knowledge on how operations in VIMRL interact with each other. During search, new instructions to be added to a program were sampled from a set of all possible programs according to probability estimates computed from the ground truth programs. These probabilities were represented as a Markov-chain where the probability of adding an instruction a program was conditioned on instruction that precedes it.

Having just a corpus of 150 ground truth programs, however, meant that the models we built were not well informed to generate programs that were successful on problems from the ARC.

\section{Results}
We ran several configurations of our system on problems from the public ARC dataset. While running these we varied the various configuration options such as the various pruning options and how final programs were selected. Our best performing configuration had all nodes with repeated inputs removed, reference gaps of greater than 2 pruned, and final choices made from unique outputs. With this configuration it was possible to solve 111 items out of the 400 public training items, and on the public evaluation side, 45 tasks were solved. Our experiments were executed on a single workstation powered by a Ryzen Threadripper 5000 CPU.

Figure \ref{fig:function-distribution} shows the distribution of functions used by our solver when generating programs for solving tasks. This chart shows how the most used operations are those for reasoning about objects.

\begin{figure}[h]
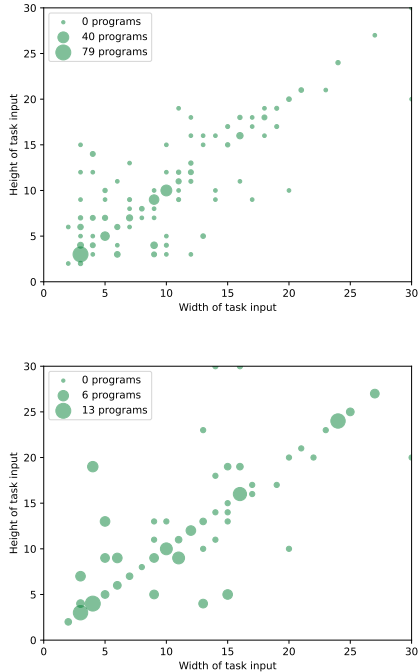

    \centering
    \includegraphics[width=0.75\linewidth]{training\_sizes.pdf}
    \includegraphics[width=0.75\linewidth]{evaluation\_sizes.pdf}
    \caption{Distribution of input grid sizes for all the tasks solved by our VIMRL based system. The size distribution for all training items solved is on the top chart, while the bottom chart shows input sizes for the evaluation items solved.}
    \label{fig:sizedistribution}
\end{figure}

Also, explicitly reasoning about objects means our system may not be limited by the sizes of the solutions it can attempt. We can see from \ref{fig:sizedistribution} how well tasks solved by the agent are spread across the spectrum of sizes tasks on the ARC occupy.




\section{Related Work}

ARC is a relatively new test, and with its development still in progress, not much work has gone into its verification. Currently, the only known human tests on the ARC are trials performed by the ARC's authors on human subjects during development \cite{Chollet2019}, and work by \citet{Johnson2021} to measure how well humans were able to infer the underlying concepts of 40 randomly selected items on the task. 

Although the scope of tasks for the study by \citet{Johnson2021} were quite limited, it showed that humans had the ability to quickly generalize the concepts behind tasks on the ARC to effectively solve them. To show how well this generalization occurred, each participant was made to provide a natural language description of their strategy which was later compared to the sequence of actions they performed while actually solving the task. In a recent work \citet{Acquaviva2021}, the authors formulated a two-player game, where one person would solve an ARC task and give a natural language description of the solution to the other player. The second player has to solve the task using the provided description. They found that at least 361 out of 400 tasks in the training set can be solved using the natural language description.


When it comes to artificial solvers, \citet{Kolev2020} present Neural Abstract Reasononer (NAR), a solver that relies on neural networks, specifically Differential Neural Computers to reason through items on the ARC. Although according to the published results, the NAR scores an accuracy of 78.8\% on items of size $10\times10$ or lower, it is not clear which section of the ARC was evaluated. \citet{fischer2020solving} developed a custom domain-specific language for the task and explored the space of programs with an evolutionary algorithm. Their approach however achieved only $3\%$ accuracy on the hidden test set.

\citet{ferre2021first} presented an approach following the minimum description length principle. The model consists of two parts: the input grid model, which converts the input grid into a parsing of the grid and the output grid model, which converts the parsing into the output grid. They tried different versions of their model, using the minimum description length principle to guide the search. Their best performing model solved 29 tasks from the training set, with each task taking upto 30 seconds.


The DreamCoder method \citet{banburski2020dreaming, alford2021neurosymbolic} presented a way of learning new abstract functions from current solved tasks and using those functions to solve more tasks in future iterations. The authors provided a set of 5 grid-manipulation to the program and selected a subset of 36 problems which could be solved using these operations. In addition, in every iteration, the method gets a fixed amount of time to solve the 36 problems. Over the iterations, the DreamCoder method is able to increase its performance from 16 tasks to 22 tasks. This is enabled because the method learns new abstract functions from the solutions in the previous iterations, thereby making the search faster in the next iteration. Though the number of problems solved by the method is low, their compression technique presents a possible way to improve search speed for solving ARC problems.

\section{Next Steps}
Our preliminary results on the ARC task show significant promise for a system that could reason generally about tasks from the ARC. The results also seem to provide some confirmation for the fact that concepts may be different across the training and evaluation sets.



The program synthesis approach with which our solver works is not significantly different from what the current best ARC solvers use. There is always a program synthesizer searching some space for possible programs. Where ours differs, however, is in the use of an imperative language that is reliant on high level functions that perform local searches. 

In the next phase of our work, we intend to expand the number and breadth of programs in our ground truth dataset; streamline operations to focus more on those that are not too specific to work on fewer tasks, and not too general to explode the search space; and improving our search algorithms.

Our current efforts have already yielded about 130 ground truth programs. Some of these programs were hand coded, while interestingly, some were discovered through our brute force searches. By expanding the ground truth dataset we have the potential to induce models that ``know'' the right way to sequence VIMRL programs. We also intend to spend some research effort on understanding the relationship between the visual appearance of the task grids and the functions we have successfully solved them with. We hope this effort will allow us to prune larger search spaces by selecting only the functions that are most likely to solve a given task.



\bibliography{main}

\end{document}